\documentclass[twoside,11pt]{article}

\usepackage{jmlr2e}

\ShortHeadings{Deep Learning for Angiodysplasia Detection and Localization}{Shvets, Iglovikov, Rakhlin, and Kalinin}
\firstpageno{1}

\usepackage[export]{adjustbox}

\begin{document}

\title{Angiodysplasia Detection and Localization Using Deep Convolutional Neural Networks}
\author{\name Alexey Shvets \email shvets@mit.edu \\
      \addr Institute for Medical Engineering and Science\\
      Massachusetts Institute of Technology\\
      Boston, MA 02142, United States of America 
      \AND
      \name Vladimir Iglovikov \email iglovikov@gmail.com \\
      \addr Lyft Inc. \\
      San Francisco, CA 94107, United States of America
      \AND
      \name Alexander Rakhlin \email rakhlin@neuromation.io \\
      \addr Neuromation OU \\
      Tallinn, 10111 Estonia
      \AND
      \name Alexandr A. Kalinin \email akalinin@umich.edu \\
      \addr Department of Computational Medicine and Bioinformatics\\
      University of Michigan\\
      Ann Arbor, MI 48109, United States of America
      } 

\maketitle

\begin{abstract}
Accurate detection and localization for angiodysplasia lesions is an important problem in early stage diagnostics of  gastrointestinal bleeding and anemia. Gold-standard for angiodysplasia detection and localization is performed using wireless capsule endoscopy. This pill-like device is able to produce thousand of high enough resolution images during one passage through gastrointestinal tract. In this paper we present our winning solution for MICCAI 2017 Endoscopic Vision SubChallenge: Angiodysplasia Detection and Localization its further improvements over the state-of-the-art results using several novel deep neural network architectures. It address the binary segmentation problem, where every pixel in an image is labeled as an angiodysplasia lesions or background. Then, we analyze connected component of each predicted mask. Based on the analysis we developed a classifier that predict angiodysplasia lesions (binary variable) and a detector for their localization (center of a component). In this setting, our approach outperforms other methods in every task subcategory for angiodysplasia detection and localization thereby providing state-of-the-art results for these problems.
% The source code for our solution is made publicly available.
The source code for our solution is made publicly available at \url{https://github.com/ternaus/angiodysplasia-segmentation}
\end{abstract}

\section{Introduction}

Angiodysplasia (AD) is the most common vascular lesion of the gastrointestinal (GI) tract in the general population \citep{foutch1995prevalence}. This condition may be asymptomatic, or it may cause gastrointestinal bleeding or and anemia \citep{regula2008vascular}. Small bowel angiodysplasia may account for 30-40\% of cases of GI bleeding of obscure origin (OGIB). In a retrospective colonoscopic analyses study, it was shown that 12.1$\%$ of 642 persons without symptoms of irritable bowel syndrome (IBS), and 11.9\% of those with IBS had colonic angiodysplasia \citep{akhtar2006organic}. In patients older than 50 years, small bowel AD is the most likely reason of OGIB \citep{sidhu2008guidelines}. \citet{liao2010indications} performed a systematic review of all original articles relevant to wireless capsule endoscopy (WCE) for the evaluation of patients with small bowel signs and symptoms published between 2000 and 2008. A total of 227 studies involving 22 840 procedures were included. OGIB (overt and occult) was the most common indication (66.0\%) and AD was the most common cause (50.0\%) of bleeding in those patients. In another study, small bowel AD lesions were the most common cause (35\%) of severe life‐threatening overt OGIB \citep{lecleire2012yield}. Lesions are often multiple, and frequently involve the cecum or ascending colon, although they can occur at other places \citep{sami2014gastrointestinal}.

\begin{figure*}[!t]
\includegraphics[width=\linewidth]{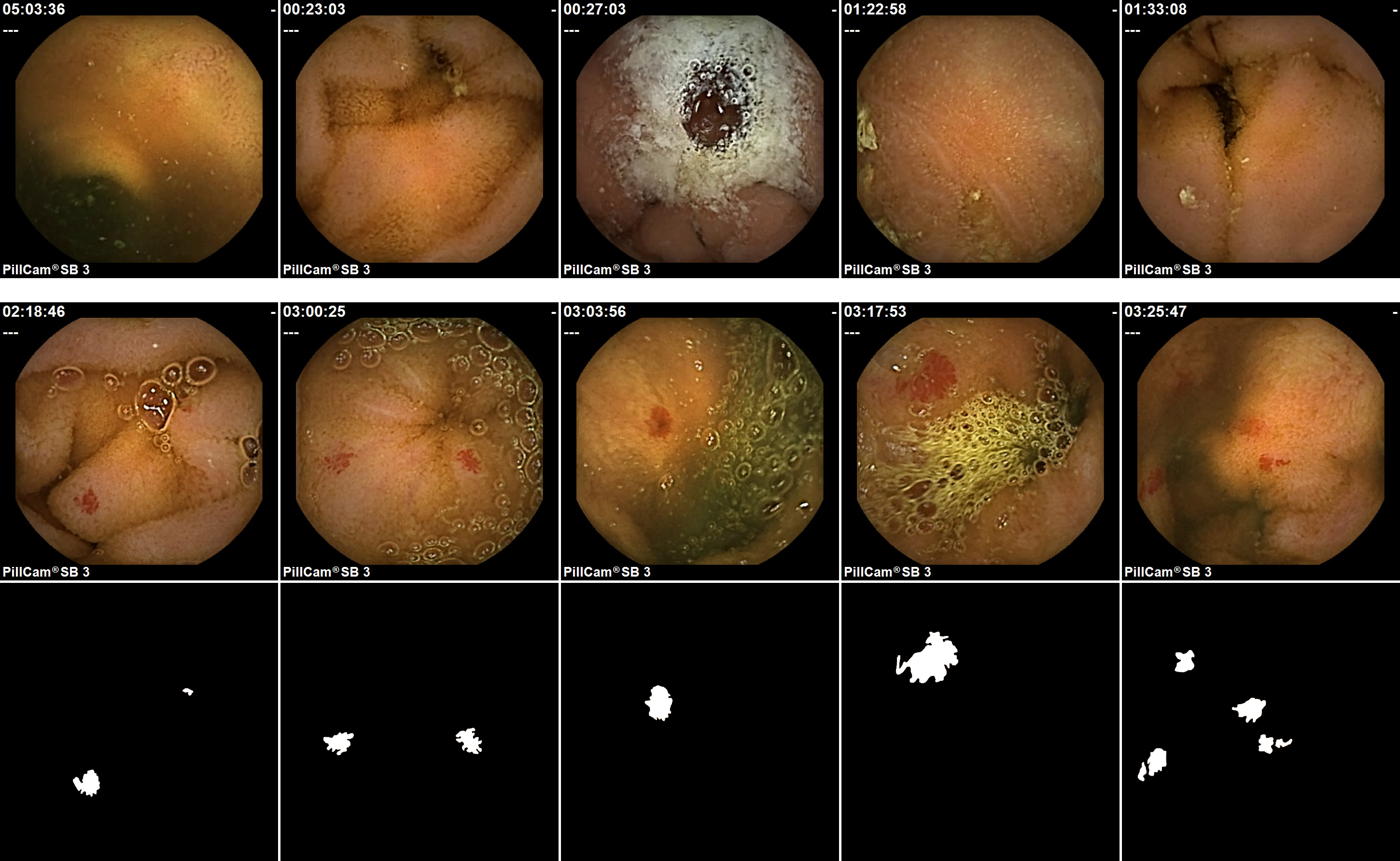}
\caption{Sample images from the training set for Angiodysplasia detection and localization challenge \citep{miccai2017}. The upper row corresponds to normal images. In the middle row the images contain angiodysplasia area represented as red spots. The down row contains masks for angiodysplasia from the middle row.}
\label{fig::dataset}
\end{figure*}

\begin{figure*}[!b]
\includegraphics[scale=2, center]{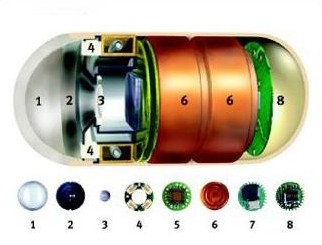}
\caption{A wireless capsule endoscope (WCE): (1) Optical Dome, (2) Lens holder,(3) Lens, (4) Illuminating LEDs, (5) CMOS imager, (6) battery, (7) ASIC RF transmitter, (8) Antenna. Source: http://www.yalemedicalgroup.org/news/ymg proctor.html}
\label{fig::wce}
\end{figure*}

The diagnosis of a vascular anomaly can be based upon endoscopic findings, histologic characteristics, or association with systemic diseases. Commonly used endoscopic modalities for assessment of the small bowel include WCE, push enteroscopy, deep small bowel enteroscopy [double‐balloon enteroscopy (DBE), single‐balloon enteroscopy (SBE) and spiral enteroscopy (SE)] or intra‐operative enteroscopy \citep{sami2014gastrointestinal}. Wireless capsule endoscopy (see Fig.\ref{fig::wce}) is the preferred first‐line investigation for the small bowel in the context of GI bleeding as it is safe, acceptable and has significantly higher or at least equivalent yield for lesions when compared with other, more invasive modalities like push enteroscopy, mesenteric angiography and intra‐operative enteroscopy \citep{triester2005meta, marmo2005meta}.
Last generation of these pill-like devices can produce more than 60 000 images with a resolution of approximately $520\times520$ pixels. However, only 69\% of angiodysplasias are detected by gastroenterologist experts during the reading of WCE videos, and blood indicator software (provided by WCE provider like Given Imaging), in the presence of angiodysplasias, presents sensitivity and specificity values of only 41\% and 67\%, respectively \citep{miccai2017}. Therefore, there is a compelling need to improve accuracy of AD detection and localization for clinical practice. In this work we apply modern deep learning techniques for automatic detection and localization of angiodysplasia.

There is a number of computer vision-based methods developed for the video capsule endoscopy analysis \citep{iakovidis2015software}, including rule-based and conventional machine learning algorithms that are applied to extracted color, texture, and other features \citep{mackiewicz2008segmentation, karargyris2010wireless, SZCZYPINSK2014segmentation}. Recently, deep learning-based approaches demonstrated performance improvements over conventional machine learning methods for many problems in biomedicine \citep{ching2017opportunities, kalinin2018deep}. In the domain of medical imaging, convolutional neural networks (CNN) have been successfully used, for example, for breast cancer histology image analysis \citep{rakhlin2018deep}, bone disease prediction \citep{tiulpin2018automatic} and age assessment \citep{iglovikov2017pediatric}, and other problems \citep{ching2017opportunities}. In the analysis of video capsule endoscopy, deep learning has recently demostrated promising results for polyp detection \citep{tajbakhsh2015automatic, yuan2017deep, byrnegutjnl2017polypsdl, murthy2017cascaded}. 

In this paper, we present a deep learning-based solution for angiodysplasia lesions segmentation from video capsule endoscopy that achieves state-of-the-art results in both binary and multi-class setting. We used this method to produce a submission to the MICCAI 2017 Endoscopic Vision SubChallenge: Angiodysplasia detection and localization \citep{miccai2017} that placed first, winning the competition. Here we describe the details of that solution based on a modification of the U-Net model \citep{ronneberger2015u, iglovikov2017satellite}. Moreover, we provide further improvements over this solution utilizing recent deep architectures: TernausNet \citep{iglovikov2018ternausnet} and AlbuNet \citep{shvets2018automatic}. To our knowledge this is first paper that try to examine angiodysplasia detection and localization using deep learning tools as a results it will serve as state of the art for other investigators.

\section{Methods}
\subsection{Dataset description and preprocessing}

A wireless capsule endoscope, is a disposable plastic capsule that weights 3.7g and measures 11mm in diameter and 26mm in length, Fig.\ref{fig::wce}. Image features include a 140 degree field of view, 1:8 magnification, 1 to 30mm depth of view, and a minimum size of detection of about 0.1mm. The capsule is passively propelled through the intesine by peristalsis while transmitting color images. Last generation of this device is able to acquire more than 60,000 images with a resolution of approximately $520\times520$ pixels \citep{mishkin2006asge}.

\begin{figure*}[!b]
\includegraphics[width=\linewidth]{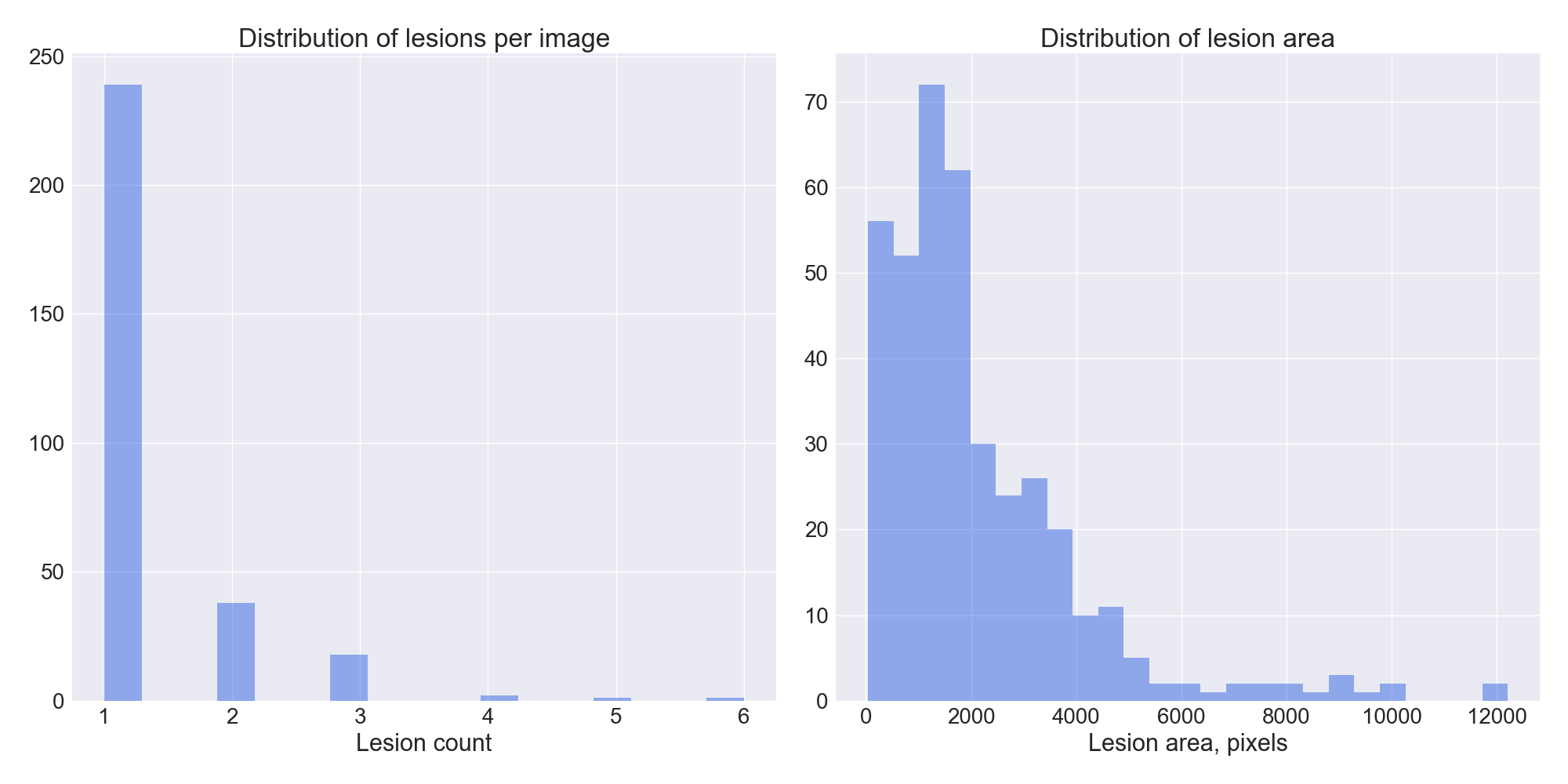}
\caption{Distribution of angiodysplasia lesions per image (left figure) and distribution of lesions area (right figure) in the data set.}
\label{fig::hist}
\end{figure*}

The dataset consists of 1200 color images obtained with WCE, Fig.\ref{fig::wce}. The images are in 24-bit PNG format, with $576\times576$ pixel resolution. The dataset is split into two equal parts, 600 images for training and 600 for evaluation. Each subset is composed of 300 images with apparent AD and 300 without any pathology. The training subset is annotated by human expert and contains 300 binary masks in JPEG format of the same $576\times576$ pixel resolution. White pixels in the masks correspond to lesion localization. Several examples from the training set are given in Fig.\ref{fig::dataset}, where the first row corresponds to images without pathology, the second one to images with several AD lesions in every image, and the last row contains masks that correspond to the pathology images from the second row. In the dataset each image contains up to 6 lesions and their distribution is shown in Fig.\ref{fig::hist} (left). As shown, the most images contain only 1 lesion. In addition, Fig.\ref{fig::hist} (right) shows distribution of AD lesion areas that reach the maximum of approximately 12,000 pixels with the median value of 1,648 pixels.

Images are cropped from $576\times576$ to $512\times512$ pixels to remove the canvas and text annotations. Then we rescale the data from $[0...255$] to $[0...1$] and standardize it following the ImageNet scheme \citep{iglovikov2018ternausnet}. For training and cross-validation we only use 299 images annotated with binary masks that contain pathology. With those, we randomly split the dataset into five folds of 60, 60, 60, 60, and 59 images. In order to improve model generalization during training, random affine transformations and color augmentations in HSV space are applied.

\subsection{Model architecture and training}
In this work we evaluate 4 different deep architectures for segmentation: U-Net \citep{ronneberger2015u, iglovikov2017satellite}, 2 modifications of TernausNet \citep{iglovikov2018ternausnet}, and AlbuNet34, a modifications of LinkedNet \citep{chaurasia2017linknet, shvets2018automatic}.

\begin{figure*}[!t]
\includegraphics[width=\linewidth]{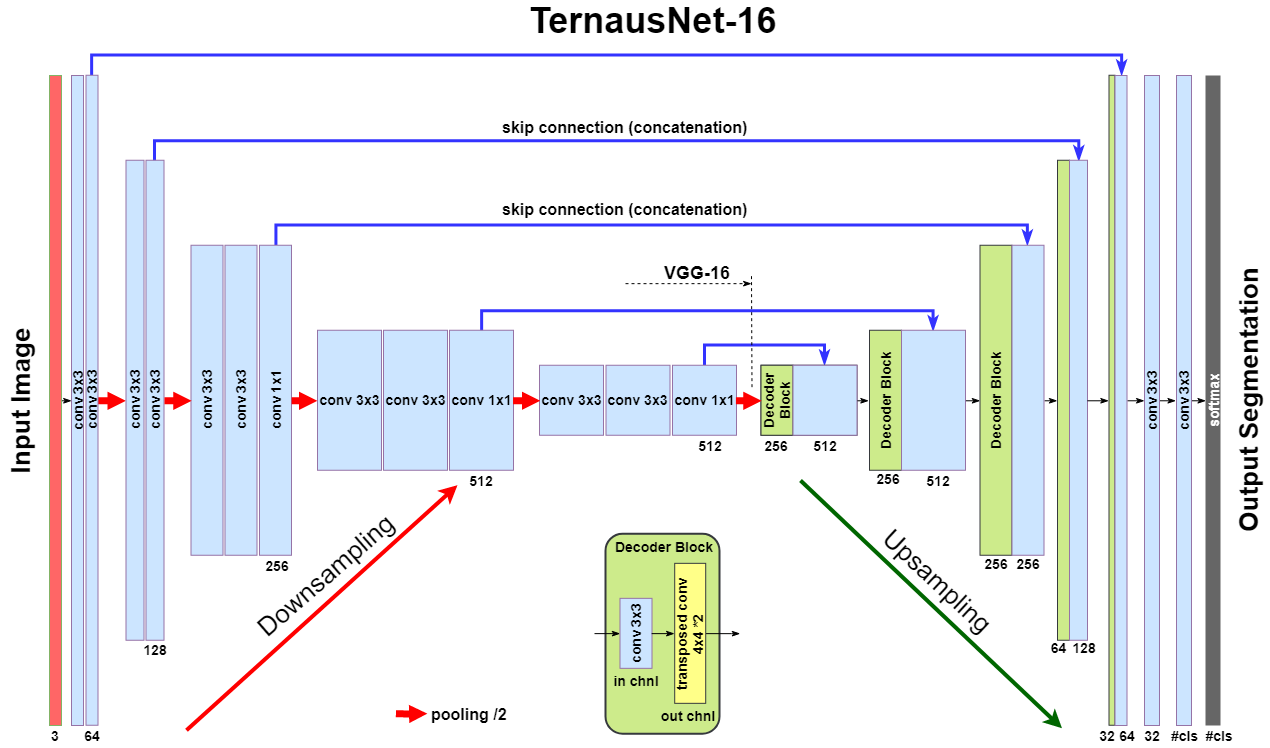}
\caption{Segmentation networks based on encoder-decoder architecture of U-Net family. TernausNet uses pre-trained VGG16 network as an encoder Each box corresponds to a multi-channel feature map. The number of channels is pointed below the box. The height of the box represents a feature map resolution. The blue arrows denote skip-connections where information is transmitted from the encoder to the decoder.}
\label{fig::ternausnet}
\end{figure*}

In general, a U-Net-like architecture consists of a contracting path to capture context and of a symmetrically expanding path that enables precise localization (for example, see Fig.\ref{fig::ternausnet}). The contracting path follows the typical architecture of a convolutional network with alternating convolution and pooling operations and progressively downsamples feature maps, increasing the number of feature maps per layer at the same time. Every step in the expansive path consists of an upsampling of the feature map followed by a convolution. Hence, the expansive branch increases the resolution of the output. In order to localize, upsampled features, the expansive path combines them with high-resolution features from the contracting path via skip-connections \citep{ronneberger2015u}. The output of the model is a pixel-by-pixel mask that shows the class of each pixel. We use slightly modified version of the original U-Net model that previously proved itself very useful for segmentation problems with limited amounts of data, for example, see \citep{iglovikov2017satellite, iglovikov2017pediatric}. Our winning submission to the MICCAI 2017 Endoscopic Vision SubChallenge: Angiodysplasia detection and localization \citep{miccai2017} was produced using this architecture.

\begin{figure*}[!b]
\includegraphics[width=\linewidth]{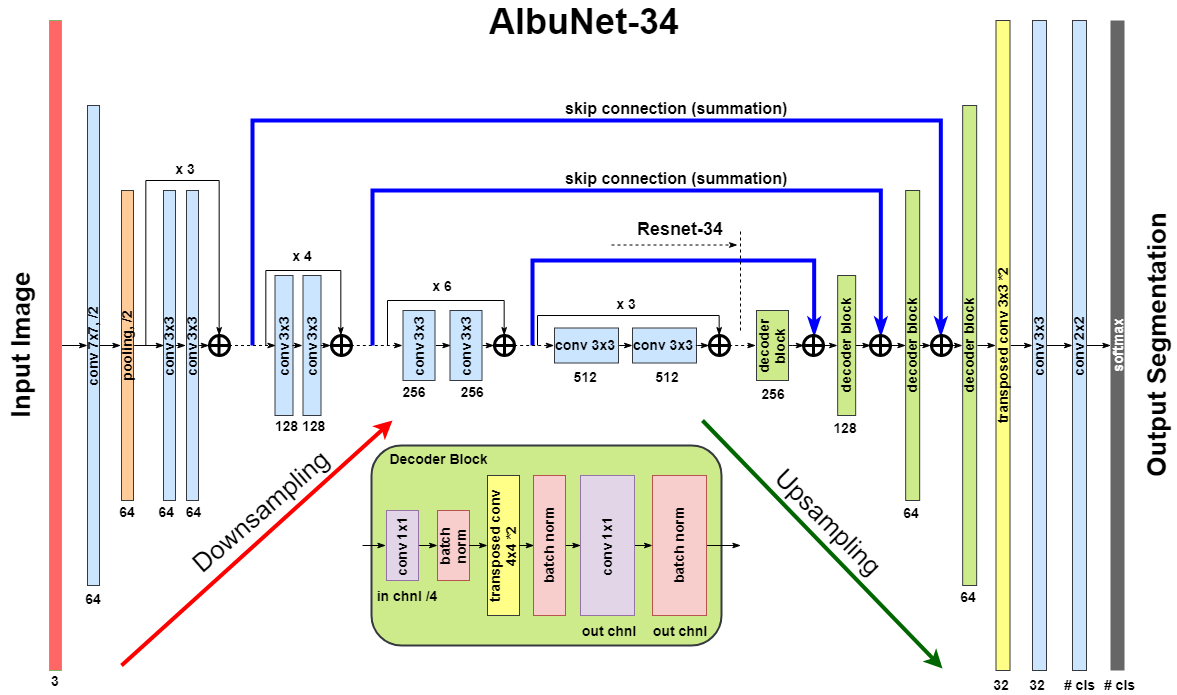}
\caption{AlbuNet-34 uses pre-trained ResNet-34 as an encoder. It is different from TernausNet in that is adds skip-connections to the upsampling path, while TernausNet concatenates downsampled layers with the upsampling path (just like original U-Net does)}
\label{fig::albunet34}
\end{figure*}

As an improvement over the standard U-Net architecture, we use similar networks with pre-trained encoders. TernausNet \citep{iglovikov2018ternausnet} is a U-Net-like architecture that uses relatively simple pre-trained VGG11 or VGG16 \citep{simonyan2014vgg} networks as an encoder (see Fig. \ref{fig::ternausnet}). VGG11 consists of seven convolutional layers, each followed by a ReLU activation function, and five max polling operations, each reducing feature map by $2$. All convolutional layers have $3\times3$ kernels. TernausNet16 has a similar structure and uses VGG16 network as an encoder (see Fig.\ref{fig::ternausnet}).

In contrast, AlbuNet uses an encoder based on a ResNet-type architecture \citep{he2016resnet}. In this work, we use pre-trained ResNet34, see Fig.\ref{fig::albunet34}. The encoder starts with the initial block that performs convolution with a kernel of size $7\times7$ and stride $2$. This block is followed by max-pooling with stride $2$. The later portion of the network consists of repetitive residual blocks. In every residual block, the first convolution operation is implemented with stride $2$ to provide downsampling, while the rest convolution operations use stride $1$. In addition, the decoder of the network consists of several decoder blocks that are connected with the corresponding encoder block. In this case, the transmitted block from the encoder is added to the corresponding decoder block. Each decoder block includes $1\times1$ convolution operation that reduces the number of filters by $4$, followed by batch normalization and transposed convolution to upsample the feature map.

We use Jaccard index (Intersection Over Union) as the evaluation metric. It can be interpreted as a similarity measure between a finite number of sets. For two sets $A$ and $B$, it can be defined as following:
\begin{equation}
\label{jaccard_iou}
    J(A, B) = \frac{|A\cap B|}{|A\cup B|} = \frac{|A\cap B|}{|A|+|B|-|A\cap B|}
\end{equation}
Since an image consists of pixels, the last expression can be adapted for discrete objects in the following way:
\begin{equation}
\label{dicrjacc}
J=\frac{1}{n}\sum\limits_{i=1}^n\left(\frac{y_i\hat{y}_i}{y_{i}+\hat{y}_i-y_i\hat{y}_i}\right)
\end{equation}
where $y_i$ and $\hat{y}_i$ are a binary value (label) and a predicted probability for the pixel $i$, correspondingly.

Since image segmentation task can also be considered as a pixel classification problem, we additionally use common classification loss functions, denoted as $H$. For a binary segmentation problem $H$ is a binary cross entropy, while for a multi-class segmentation problem $H$ is a categorical cross entropy.

The final expression for the generalized loss function is obtained by combining (\ref{dicrjacc}) and $H$ as following:
\begin{equation}
\label{free_en}
L=H-\log J
\end{equation}
By minimizing this loss function, we simultaneously maximize probabilities for right pixels to be predicted and maximize the intersection $J$ between masks and corresponding predictions. We refer reader to \citet{iglovikov2017satellite} for further details. Each model is trained with Adam optimizer \citep{DBLP:journals/corr/KingmaB14} for 10 epochs with learning rate 0.001, and then for another 5 epochs with the learning rate 0.0001.

As an output of a model, we obtain an image, in which each pixel value corresponds to a probability of belonging to the area of interest or a class. The size of the output image matches the input image size. For binary segmentation, we use $0.3$ as a threshold value (chosen using validation dataset) to binarize pixel probabilities. All pixel values below the specified threshold are set to $0$, while all values above the threshold are set to $255$ to produce final prediction mask.

\begin{figure*}[!b]
\includegraphics[width=\textwidth]{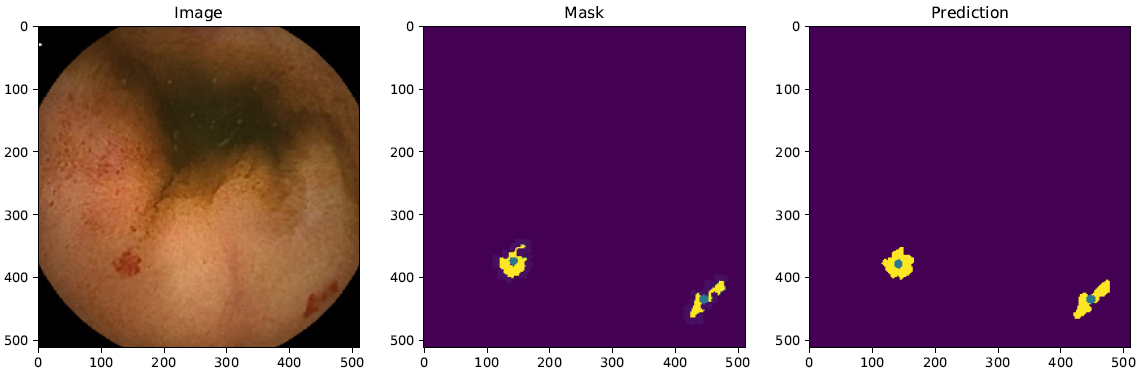}
\caption{The prediction of our detector on the validation data image. Here, the first picture correspond to original image, the second one to the training mask, the last one to the predicted mask. Green dots inside the clusters corresponds to the centroid coordinates that define a localization of the appropriate angiodysplasia. For example, the real values for centroid coordinates are $p^1_{mask}$ = (376, 144), $p^1_{pred}$ = (380, 143) for the first cluster and $p^2_{mask}$ = (437, 445), $p^2_{pred}$ = (437, 447) for the second one.}
\label{fig::train_angio}
\end{figure*}

Following the segmentation step, we perform postprocessing in order to find the coordinates of angiodysplasia lesions in the image. In the postprocessing step we use OpenCV implementation of connected component labeling function: connectedComponentsWithStats \citep{bradski2000opencv}. This function returns the number of connected components, their sizes (areas), and centroid coordinates of the corresponding connected component. In our detector we use another threshold to neglect all clusters with the size smaller than 300 pixels. Therefore, in order to establish the presence of the lesions, the number of found components should be higher than 0, otherwise the image corresponds to a normal condition. Then, for localization of angiodysplasia lesions we return centroid coordinates of all connected components. 

\section{Results}
To test our prediction and compare it with known mask we performed calculations on an image taken from the validation set.
The exemplar result of the prediction is shown in Fig.\ref{fig::train_angio}. For a visual comparison we also provide the original image and its corresponding mask. Given imperfect segmentation, this example does show that the algorithm sucessfully detects angiodysplasia lesions. When there are few lesions in an image and they are well separated in space, the detector performs almost very well. In case of many lesions that somehow overlap in space, further improvements are required, specifically in choosing model hyperparameters, to achieve better performance.

% the same as for robot
The quantitative comparison of our models' performance is presented in the Table\ref{table:segmentation}. For the segmentation task the best results is achieved by AlbuNet-34 providing $IoU = 0.754$ and $Dice = 0.831$. When compared by the inference time, AlbuNet is also the fastest model due to the light encoder. In the segmentation task this network takes around 20$ms$ for $512\times512$ pixel image and more than three times as fast as TernausNets. The inference time was measured using one NVIDIA GTX 1080Ti GPU.

% In such a way our models can provide a comparable performance with the results obtained for the binary instrument segmentation.

\begin{table}[t!]
\caption{Segmentation results. Intersection over Union (IoU) and Dice coefficient (Dice) are in $\%$ and inference time (Time) is in $ms$.}
\label{table:segmentation}
\centering   
\begin{tabular}{|c | c | c | c |}
\hline
Model & IOU & Dice & Time \\
\hline 		
U-Net         &          73.18 &          83.06 &         30 \\
TernausNet-11 &          74.94 &          84.43 &         51 \\
TernausNet-16 &          73.83 &          83.05 &         60 \\
AlbuNet-34    & \textbf{75.35} & \textbf{84.98} & \textbf{21} \\
\hline
\end{tabular}
\end{table}

\section{Conclusions} 
We present deep learning-based segmentation and detection algorithms for angiodysplasia lesions localization in video capsule endoscopy. This study compares U-Net network architecture with its improved modifications that use custom pretrained encoders. Subsequent postprocessing based on the analysis of connected components is used to further refine predictions. Our approach shows quite good results on the validation. To the best of our knowledge, this study presents the first attempt in appllication of convolution neural networks for the problem of angiodysplasia lesion detection and classification. We demostrate state-of-the-art results in the MICCAI 2017 Endoscopic Vision SubChallenge: Angiodysplasia Detection and Localization. These results can be further improved by more accurate  hyperparameter tuning as well as better postprocessing of connected components.
% The source code for our approach is made publicly available under MIT licence.
Our code is available as an open source project under MIT licence at \url{https://github.com/ternaus/angiodysplasia-segmentation}.

%This is where you give us some technical understanding about why your awesomeness is awesome, and maybe even some times when it isn't so we know when we should be using it.  Discuss both technical and clinical implications, as appropriate.  

%Make sure you also put your awesomeness in the context of related work.  Who else has worked on this problem, and how did they approach it?  What makes your direction interesting or distinct?

% ACKNOWLEDGEMENTS ONLY GO IN THE CAMERA-READY, NOT THE SUBMISSION
% \acks{Many thanks to all collaborators and funders!}

\bibliography{paper}

\end{document}